# A STACKED AUTOENCODER NEURAL NETWORK BASED AUTOMATED FEATURE EXTRACTION METHOD FOR ANOMALY DETECTION IN ON-LINE CONDITION MONITORING

**Mohendra Roy**
School of Electrical and Electronic Engineering
Nanyang Technological University
Singapore 639798
mohendra.roy@ntu.edu.sg *

**Sumon Kumar Bose**
School of Electrical and Electronic Engineering
Nanyang Technological University
Singapore 639798
Bose0003@e.ntu.edu.sg

**Bapi Kar**
School of Electrical and Electronic Engineering
Nanyang Technological University
Singapore 639798
bapik@ntu.edu.sg

**Pradeep Kumar Gopalakrishnan**
School of Electrical and Electronic Engineering
Nanyang Technological University
Singapore 639798
pradeepgk@ntu.edu.sg

**Arindam Basu**
School of Electrical and Electronic Engineering
Nanyang Technological University
Singapore 639798
arindam.basu@ntu.edu.sg

## ABSTRACT

Condition monitoring is one of the routine tasks in all major process industries. The mechanical parts such as a motor, gear, bearings are the major components of a process industry and any fault in them may cause a total shutdown of the whole process, which may result in serious losses. Therefore, it is very crucial to predict any approaching defects before its occurrence. Several methods exist for this purpose and many research are being carried out for better and efficient models. However, most of them are based on the processing of raw sensor signals, which is tedious and expensive. Recently, there has been an increase in the feature based condition monitoring, where only the useful features are extracted from the raw signals and interpreted for the prediction of the fault. Most of these are handcrafted features, where these are manually obtained based on the nature of the raw data. This of course requires the prior knowledge of the nature of data and related processes. This limits the feature extraction process. However, recent development in the autoencoder based feature extraction method provides an alternative to the traditional handcrafted approaches; however, they have mostly been confined in the area of image and audio processing. In this work, we have developed an automated feature extraction method for on-line condition monitoring based on the stack of the traditional autoencoder and an on-line sequential extreme learning machine(OSELM) network. The performance of this method is comparable to that of the traditional feature extraction approaches. The method can achieve 100% detection accuracy for determining the bearing health states of NASA bearing dataset. The simple design of this method is promising for the easy hardware implementation of Internet of Things(IoT) based prognostics solutions.

# 1   Introduction

Condition Monitoring(CM) of machine health and maintenance are parts of the routine job in all major process industries. This is also an integral part of the evaluation of risk and asset management [1]. Specially, the on-line monitoring and diagnosis of machine health help in quantifying the impact of aging parts on the reliability of the overall system. Further, this helps in improving system safety, reduce processing and operation time, increase system availability, maintenance planning, alerting the crew about the impending failure and from being reactive to be proactive, etc. [2]. Recently there has been an increase in on-line machine health monitoring research aiming for predictive maintenance and to extend equipment's service life, as well as to develop on-board Integrated Systems Health Management (ISHM). This has been fuelled by the recent development in the machine learning research [3, 4, 5, 6, 7]. Many of them are possible due to the increase in the computation capability, efficient models as well as the availability of dataset. To further accelerate this type of research, the National Aeronautics and Space Administration (NASA) in conjunction with various industry and academia has open sourced several datasets related to machine health [8].

Many research groups have successfully developed various prognostics models based on these datasets [9]. Most of these datasets are generated by acquiring the raw sensor signals. This increases the dimension of the dataset and may also contain unwanted interference and noise. The dimensionality of the data has a direct impact on the accuracy as well as training time of the artificial neural network based models, and it turns out to be more crucial for the applications like edge computing and Internet of things(IoT) based diagnosis solutions [10]. A popular approach to solving this kind of problem is to reduce the dimension of the raw input signals, which streamlines the training and utilization times, data visualization, data understanding, reduce the memory requirement, etc. Several methods are there for achieving this [11]. For example, extracting the features by taking (a) root mean square(RMS) of the raw feature vector [12], (b) Kurtosis [13, 14], (c) Skewness [15, 16], (d) Crest factor [15, 16], (e) Peak to Peak [17, 18], and etc.

All these are handcrafted approaches and require domain knowledge combined with "trial and error" strategies. However, recent progress in autoencoder based feature extraction has provided an alternative means for feature extraction [19, 20]. The autoencoder based feature extraction is not only helped in solving the curse of dimensionality [21] but also can provide more discriminative features compared to the traditional feature engineering approaches [22]. Moreover, the best features for that task are directly "learnt" from the data avoiding adhoc trial and error strategies. In some cases, it outperforms the traditional handcrafted features [23]. A considerable amount of progress has been made in terms of automated image and speech feature extraction [24]. However, so far its application in terms of condition monitoring is lacking.

In this paper, we have demonstrated a stack of the traditional autoencoder(TAE) and an On-line Sequential Extreme Learning Machine(OSELM) for automated feature extraction and condition monitoring of bearing health. The objective of this work is to develop an automated feature extraction method followed by an on-line condition monitoring system,
which is comparable to the traditional feature extraction approaches. The detail of this method and results are described
in the following sections.

# 2   Methods

## 2.1   Stack of artificial neural networks

For the automated feature extraction, we have used the traditional autoencoder[25, 26, 27], which was trained to find the compress representation of the input data. The hidden layer output from this trained autoencoder was then used as an input to the boundary type OSELM [28] (see figure 1) for the inference of the bearing health.

The traditional autoencoder is an artificial neural network that can be trained to produce a replica of its input to its output [29]. The reduced code layer $h$ can capture the compressed representation of the input. This compressed representation can be used as extracted features. An autoencoder consists of two parts; the encoder part takes the input $X \epsilon \mathbb{R}^d$ and maps it to the $h \epsilon \mathbb{R}^L$, where

$$h = \sigma \left( WX + b \right) \tag{1}$$

This $h$ is generally referred to as code layer or latent representation. Here $\sigma$, $W$ and $b$ are the activation function, input weight matrix, and bias vector respectively. The decoder part maps the latent representation $h \epsilon \mathbb{R}^L$ to the output $\hat{X} \epsilon \mathbb{R}^m$ (for autoencoder $m = d$). Here $\hat{X} = \gamma \left( W^0 h + b^0 \right)$, where $\hat{X}$ is the reconstruction of the input and $\gamma$, $W^0$, $b^0$ are the activation function, output weight matrix, and bias vector respectively. In this work, we have used a single hidden layer autoencoder with $d = 4096$, $L = 5$ and $m = 4096$. We have used the rectified linear units(ReLU) as an activation function in both encoder and decoder parts [30, 31] and the weights were optimized using the Adam optimizer [32].

The network was first trained in off-line mode using the training dataset for one epoch, keeping out the test dataset for the on-line feature extraction. The trained system (only the encoder part) was then used for the on-line feature extraction for the raw samples (see Figure 1) in the testing mode. Here the samples from the test dataset were fed into the input layer of the autoencoder(see network B in Figure 1) in sequence to compute the hidden layer output(the hidden layer output of network B is the same to the input of network C in Figure 1), and in each instance, these hidden layer outputs are then fed to the input layer of the OSELM (see the input of network C in Figure 1) for the on-line training of OSELM and subsequently for the inference (i.e., condition monitoring). To summarize, the encoding layer of the autoencoder is learnt offline and can be deployed on the condition monitoring hardware platform as a feature extractor.

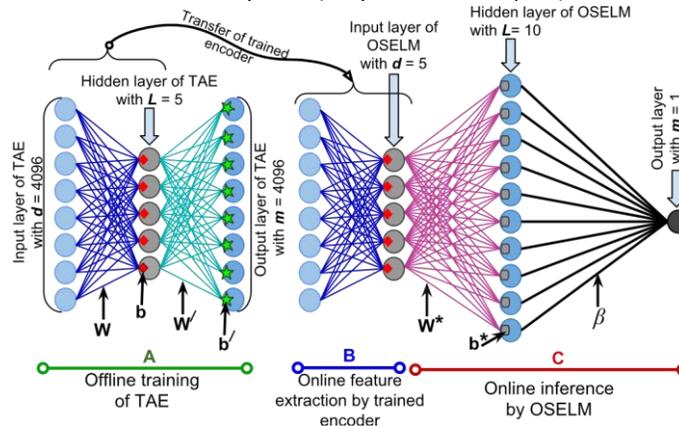

Figure 1: Graphical illustration of the stack arrangement of neural networks for the on-line feature extraction and condition monitoring. This stack arrangement is consisting of two types of neural networks. The $1^{st}$ type is of a traditional autoencoder(TAE) having the input and output layers of same dimension, i.e. of 4096 neurons and a single hidden layer with five neurons (see network $A$). The $2^{nd}$ type of network (network $C$) is of on-line sequential extreme learning machine(OSELM), having an input layers with 5 neurons, hidden layers with 10 neurons, and a single output node. In our proposed method the network $A$ was first trained in off-line mode using the training dataset with smoothed input data to learn to get the compressed representation of the data. This trained network (network $B$) was then used (i.e. only the encoder part of network $A$) for the on-line extraction of the features from the test dataset with raw features. This extracted features from the trained encoder (i.e. output of network $B$) were then fed into the classifier network $C$ (input layer of OSELM), and used for on-line learning. After the convergence, this trained OSELM was then used for the inference of the rest of the on-line samples.

The OSELM is an on-line version of Extreme Learning Machine(ELM), which is known for its fast and straightforward learning method [28, 33]. Typically, an ELM is a single hidden layer feed forward network, where the input

weights($W^*$) and biases($b^*$) are generated randomly from a continuous probability distribution and kept intact without updating and the hidden layer outputs are computed as in equation 1. Only the output weights of the network are usually learnt in a single step resulting in quicker convergence. In batch mode, the output weights are determined as:

$$\beta = H^T \left( \frac{I}{C} + HH^T \right)^{-1} Y$$

(2)

Here $H$, $I$, $Y$ and $C$ are the hidden layer output, identity matrix, target output and a hyperparameter respectively. For the on-line version of ELM, the $\beta$ is updated sequentially such as:

$$\beta_n = \beta_{n-1} + P_n^{-1} \left( Y_n - H_n \beta_{n-1} \right) H_n^T$$

(3)

Here $P_n = \left( P_{n-1} + H_n H_n^T \right)$ and for the 1st batch of samples, i.e., for n = 1, $P_{n-1} = P_0 = \left( \frac{I}{C} + H_0 H_0^T \right)$ [34].

Here we have used the OSELM as a one class classifier or anomaly detector in boundary mode [35, 36]. That means the network was trained for only one particular class (with only healthy samples from the machine assuming health condition has not degraded at the beginning of the lifetime) and the output of the hidden layer was mapped to only one output node ($Y = 1$ for healthy samples). As a result, it required less number of multiplication and accumulation(MAC) operation compared to that of the reconstruction mode where the number of output neurons is same as input dimension $d$. The input layer of the OSELM has the same dimension as that of the hidden layer of the traditional autoencoder (see figure 1). The hidden layer (of OSELM) is having a dimension of $L = 10$ and a single output node, i.e., $m = 1$(see network C in figure 1). We have trained this network in an on-line fashion with features that were extracted by the traditional autoencoder (i.e., the trained encoder part of the autoencoder, see figure 1). The initial $\beta_0$ was calculated using the first 10 samples in a batch. Then the $\beta$ was kept updating on-line, i.e., sample by sample until it got converged. Thus the network was trained by the on-line extracted features. Since the bearings are considered to be in healthy state at the earlier stage of their life, therefore the network got trained only for the healthy class. The converged network was then used for the on-line inference for rest of the samples. In this mode, if the samples were from the unhealthy state of the bearing, then the output of the network got deviated from its trained class, which resulted in high difference from $Y = 1$ at the output. Any difference more significant than the defined threshold was considered as an anomaly or faulty state. Thus we obtained the condition of the bearing health.

Since the algorithm converges very fast [37], therefore we can train it using a few samples. The converge criteria is based on how small the change of $\beta$ has become during the update (or training), i.e., $\Delta\beta$ should be very small for a fair number of consecutive samples. In our method we used %$\Delta\beta < Tc$ for a consecutive 10 samples as a convergence criterion. Here $Tc = 0.1\%$ is the termination value. The average convergence length for the dataset with auto extracted features is 582, which is about 5 times less to that of the average sample length of the dataset (the average sample length of the NASA dataset is 3154). For the same dataset, the average convergence length is 481 with handcrafted features. This shows that for the both type of feature extraction, the network gets converged within the first $(1/5)^{th}$ of the total samples of the dataset. Thus OSELM is suitable for the on-line training and subsequently for inference using the same dataset. From the viewpoint of the application, a different model has to be learnt for each machine since the placement of sensors and type of fault of each machine may be unique. Hence, it is best if healthy models are learnt online for each machine separately in an online manner on the deployed hardware. The choice of ELM as the classifier is motivated by both its fast convergence as well as availability of power efficient hardware [38, 39] for deployment.

## 2.2   Preparation of dataset for on-line feature extraction

In this work we have used the NASA Bearing dataset [8]. This dataset was provided by the Center for Intelligent Maintenance Systems (IMS), University of Cincinnati. This contains three types of datasets, each of them consisting of vibration signals from four bearings. The dataset 1 contains vibration signals for all the four bearings which were acquired from the accelerometers in both the X and Y axis. For the dataset 2 and 3, the vibration signals for all the four bearings were obtained only from the accelerometer in the X-axis. These vibration signals were recorded for the duration of 1 sec at an interval of 10 minutes with a sampling rate of $20KHz$. The description of each bearing dataset and their health status are as given in table 1.

All these are the raw signals, which also carry some amount of noise. We have pre-processed these raw signals to remove some of these noise by taking the average value of every consecutive five raw data samples. This averaging helps in eliminating high-frequency noise. Also, it reduces the number of features in every sample to $20480/5 = 4096$, which is also helpful in reducing the computation overhead and training time.

We have used these filtered signals for rest of this work, i.e., for the training of the network as well as for the evaluation of bearing health. For this purpose, we have arranged the whole NASA bearing dataset into two parts, as training and testing dataset using leave one out strategy [40].

Since the NASA bearing dataset has total 12 bearings (see table 1), therefore we have arranged 12 sets of datasets. Each set contains 11 bearing datasets for training and the left out one for the testing. For example, to evaluate the condition of the bearing1 of dataset1, we have treated the Dataset1_Bearing1 as test dataset (i.e. for on-line testing) and rest of the remaining 11 bearing datasets as training dataset (i.e. in off-line training mode).

## 2.3   Preparation of dataset with handcrafted features

For the comparison of our proposed method, we have performed the same on-line inference using five handcrafted features extracted from the raw data. We have used the following five time-domain methods to create the handcrafted features: RMS, kurtosis, skewness, crest factor and peak to peak. We have calculated each of these parameters from the raw features for all the samples. Thus we have the dataset of 12 bearings with five handcrafted features. Since in this case, we are not going to extract the features using the autoencoder, therefore we don't have to train it. Here we are only doing the on-line evaluation of the bearing condition by using only the OSELM network (see the network C in figure 1). For that, we fed the data from the dataset with handcrafted features in the input layer of the OSELM (sample by sample) and obtained the on-line inference after it converged. The method of training and inference of the OSELM network is as described in the previous section.

# 3   Results and discussion

To check the reliability of the proposed method, we have obtained the maximum of the deviation of the OSELM output to that of the healthy state for the entire lifetime of the bearing (i.e. length of the dataset). Figure 2(a) shows the maximum deviation of the OSELM output from that of the healthy state for each of the bearing, using the automatically extracted features from the autoencoder. From the figure we can see that the Dataset1_Bearing3, Dataset1_Bearing4, Dataset2_Bearing1, and Dataset3_Bearing3 are having comparatively more deviation than the rest of the bearings. This is because all these four are the faulty bearings according to the ground truth (see the dataset description in table 1). The results from the handcrafted features also follow a similar pattern (see figure 2(b)). This shows that the proposed method is able to indicate the bearing health status. However, we want to develop an automated process that spontaneously indicate the condition of the bearing health. This requires a scheme to choose an appropriate threshold value. Therefore, we have implemented an adaptive thresholding method that is based on the average value and the standard deviation of the all the output deviations of OSELM at the time of on-line training. These deviations are a measure of the amount of noise in the data. The threshold $T$ is given by:

$$T = K\left(\mu_t + \sigma_t\right) \qquad (4)$$

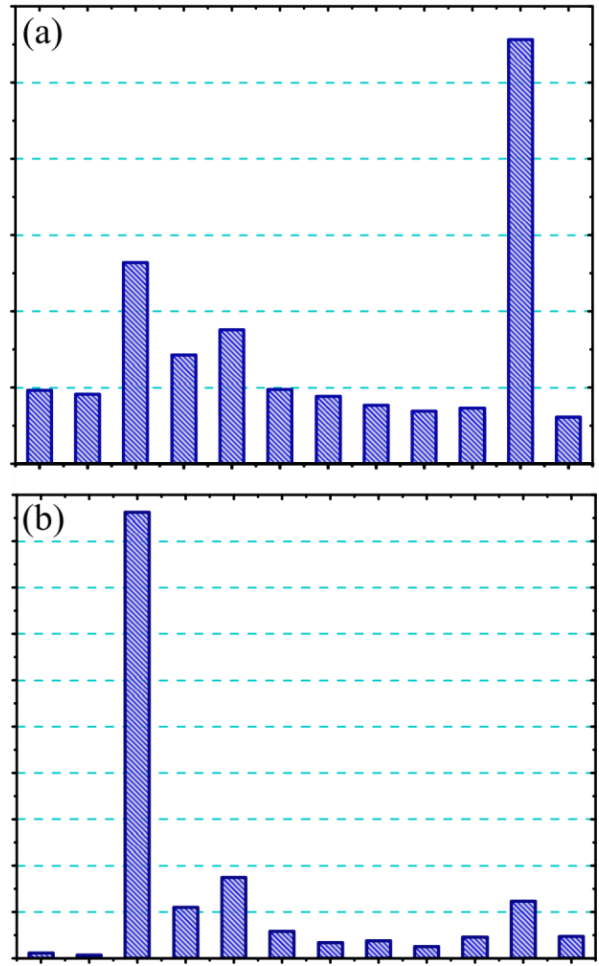

Figure 2: Condition monitoring result of NASA bearing dataset using both the proposed feature extraction method and traditional handcrafted approach. Maximum deviation from expected value is shown for each test bearing data for (a) automatically extracted features and (b) handcrafted features. In both the figures the faulty Data1_Bearing3, Data1_Bearing4, Data2_Bearing1, and Data3_Bearing3 are having more deviation compared to the rest of the good bearings, which correlate well with the dataset description as in table 1.

Table 1: Detailed description of NASA bearing dataset [8]

| Dataset | Bearing | No. of Samples | No of features in each sample | Condition |
|---------|---------|----------------|-------------------------------|-----------|
| Dataset1 | Bearing1 | 2156 | 20480 | No defect |
| | Bearing2 | 2156 | 20480 | No defect |
| | Bearing3 | 2156 | 20480 | Defect in inner race |
| | Bearing4 | 2156 | 20480 | Roller element defect |
| Dataset2 | Bearing1 | 984 | 20480 | Outer race failure |
| | Bearing2 | 984 | 20480 | No defect |
| | Bearing3 | 984 | 20480 | No defect |
| | Bearing4 | 984 | 20480 | No defect |
| Dataset3 | Bearing1 | 6324 | 20480 | No defect |
| | Bearing2 | 6324 | 20480 | No defect |
| | Bearing3 | 6324 | 20480 | Outer race failure |
| | Bearing4 | 6324 | 20480 | No defect |

This was calculated at the time of on-line training of the OSELM. Here $\mu_t$ is the average of all the deviations (or differences) of the training samples from the healthy class during the training, $\sigma_t$ is the standard deviation of all these deviations (i.e. the differences), and $K$ is the hyper parameter which has to be adjusted. In the inference mode, for any bearing dataset if the output deviation of the OSELM was higher than the corresponding threshold, then it was considered as the unhealthy state of the bearing.

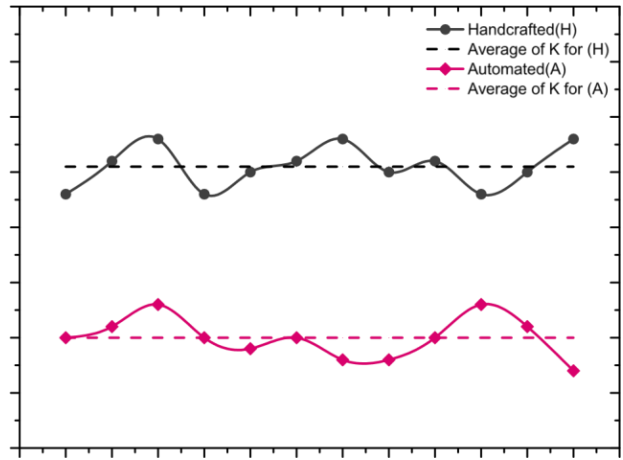

Figure 3: Evaluation of $K$ for determining the adaptive threshold for each of the test dataset. Here the plot in black is for the dataset with traditional handcrafted features and the plot in magenta is for the dataset with the automatically extracted features. In both the cases the $K$ value was evaluated using the leave one out strategy.

Since the threshold also depends on the hyper parameter $K$ therefore, we need to optimize the value of $K$ for higher accuracy (in determining the bearing health state). For that we used the same leave one out strategy. In other words, we have used the same training dataset that was used for the training of the autoencoder to find the optimized $K$ value and used it for determining the bearing state of the test dataset. After finishing the training of the autoencoder, we used the same training dataset containing the 11 bearing data. For each bearing dataset, we extracted their compressed features (using the trained autoencoder) and used it for the on-line training of the OSELM. During the training we evaluated the $\mu_t$ and $\sigma_t$ and then found the value of $K$ which determined the true state of the corresponding bearing dataset (in the inference mode). We selected the one $K$ value for which it shows the maximum accuracy in determining the bearing health sate of all the 11 bearings in the training dataset. After finding the optimum value of $K$, we then used it for the actual test dataset (i.e. which was left out as a test dataset) and evaluated its performance in determining the true state of the bearing health as per as the ground truth (as described in the table 1). In all cases, the value of $K$ chosen from training data was able to successfully classify healthy vs faulty bearing in the test data. The optimized $K$ value that correctly detected the health sate of each of the test dataset is as shown in the figure 3. Also, we did the the same for the handcrafted dataset. The results are as shown in the figure 3.

We further validate our findings by evaluating the percentage of accuracy in determining the bearing health state of all the 12 bearings against various values of $K$ in test mode. Here the percentage of accuracy is the accuracy in detecting the bearing state of all the bearings in the context of ground truth as described in table 1. The $K$ Vs. percentage of accuracy for the auto-extracted features shows that the accuracy becomes 100% for $K \approx 10$ (see figure 4). That means predictions for all the bearings is analogous to the ground truth. For the handcrafted features, this can be achieved for $K \approx 25$(see figure 4). These values roughly match the average values of $K$ obtained from the training dataset as described earlier and explain the good performance achieved in the earlier test. Though the $K$ value of the automated method and the handcrafted method are different, but both the approaches can accurately predict the bearing states of all the bearings.

From these results, it is clear that the automatically extracted features are also as useful as handcrafted features in detecting the condition of the bearing health. Since the automated feature extraction method does not require any prior knowledge of the dataset and it requires to train only once for a particular type of dataset, therefore the proposed method is more feasible for applying to different machines.

## 4 Conclusion

In conclusion, we have demonstrated an automated feature extraction method that can extract the compressed representation of the raw signals, which can be utilized for the on-line condition monitoring. The performance of this method is similar to that of the handcrafted feature extraction approaches. Since the training of the autoencoder can be performed off-line, therefore we can extract the features on-line using the trained hyper-parameters only. This makes it very useful to implement in hardware and to make edge computation based IoT applications. Together with the on-line ELM, which is already preferred for its simplicity to implement in hardware by various research groups, the proposed method may provide an alternative IoT based condition monitoring system.

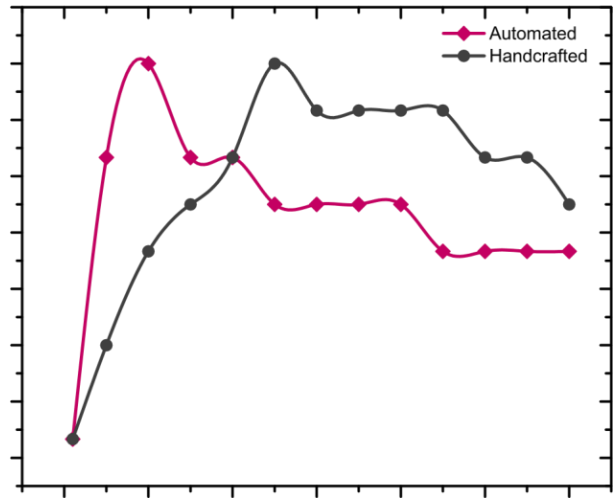

Figure 4: $K$ Vs. % of detection accuracy in determining the bearing health state of all the bearings of NASA bearing dataset as per the ground truth as described in table 1. The plot in black is for the dataset with handcrafted features and the plot in magenta is for the dataset with automatically extracted features. This result validates the result from figure 3. which shows that the for handcrafted features the optimum value of $K \approx 25$ and for automated features it is about 10. Again it shows that both the methods are capable in detecting bearing states with 100% accuracy.